# Generative AI-Based Text Generation Methods Using Pre-Trained GPT-2 Model

Capstone project report submitted in partial fulfillment of the requirements
for the Post Graduate Program in Data Science and Artificial Intelligence
at Praxis Tech School, Kolkata, INDIA

By

Rohit Pandey (A23034)
Hetvi Waghela (A23019)
Sneha Rakshit (A23043)
Aparna Rangari (A23006)
Anjali Singh (A23004)
Rahul Kumar (A23031)
Ratnadeep Ghosal (A23032)

Under the supervision of
Prof. Jaydip Sen
Praxis Tech School, Kolkata, India

# Generative AI-Based Text Generation Methods Using Pre-Trained GPT 2 Model

Rohit Pandey[1], Hetvi Waghela[2], Sneha Rakshit[3], Aparna Rangari[4], Anjali Singh[5], Rahul Kumar[6], Ratnadeep Ghoshal[7], Jaydip Sen[8]
Email: {[1]rohit.pandey_dsfall, [2]hetvi.mahendra.waghela_ds23fall, [3]sneha.rakshit_ds23fall, [4]aparna.arvind.rangari_ds23fall, [5]anjali.singh_ds23fall, [6]rahul.kumar_ds23fall, [7]ratnadeep.ghoshal_ds23fall}@praxistech.school, [8]jaydip@praxis.ac.in

## 1. Introduction

A text generation model is a machine learning model that uses neural networks, especially transformers architecture to generate contextually relevant text based on linguistic patterns learned from extensive corpora. The models are trained on a huge amount of textual data so that they can model and learn complex concepts of any language like its grammar, vocabulary, phrases, and styles.

Text generation models can increase the productivity of humans in their current business processes. These models are already automating the process of content creation across industries for the generation of reports, summaries, and emails among others. These models are also allowing for a greater level of personalization in communications between businesses and their customers. These models are also able to summarize long articles and books, making information more widely understandable and comprehensible with a shorter turnaround time. Nowhere, has the impact of text generation models been greater than in the field of machine translation. With, the large amount of textual data corpus that we have across several languages, we are seeing the long-promised ability of models to perform a variety of tasks across languages other than English come to fruition. These text generation models are most importantly being used by humans to augment themselves, through a human plus machine approach. In this approach, these models are used to generate ideas and are used by humans as a solution space reduction method as opposed to a definite solution generator. The use cases are expanding by the day allowing humans to do higher value-added tasks or do their current tasks with more efficiency.

The earliest text generation models can be traced back to the 1950s in the form of statistical methods like n-grams, which calculated the probability of the occurrence of a word, based on the previous *n*-1 words. N-gram models are simple and efficient and have a very good performance on certain tasks like autocomplete. But, because the lookback window is restricted to *n*-1 words, they have limited capability to be able to capture longer-range dependencies in text. They are also affected by the curse of dimensionality and data sparsity. With data sparsity, it becomes difficult to model low-frequency but plausible sequences of text.

The next evolution was the use of rules and statistics-based models throughout the 1980s and 1990s, where the model used word frequency and analysis of grammatical structures to generate text. The rules-based models relied on linguistic rules hand-crafted by experts. These rules were logical constructs that had defined relationships about the interplay between grammar, meaning, and use case. These rules were then used to make inferences and generate text. However, creating these rules was time-intensive and failed to capture complex relationships and the high variability of natural languages. During this period, we also had hidden Markov models, which were developed conceptually in the 1960s, however, it was the advancement of computational resources and the availability of large datasets which allowed the true potential of HMM to surface. HMMs were used for Part-of-Speech tagging, Speech Recognition, and Text generation, but even it suffered from the problem of limited context i.e. reliance on intermediate state and data sparsity issues. However, HMMs did set the stage for further advancements in the field, by demonstrating the power of statistical language models.

The next era was marked by a sea change, with the introduction of machine learning and neural networks in the late 1990s. Machine Learning Algorithms like Decision trees and Support vector machines were being used in NLP. Decision trees were most useful in POS tagging and syntactic parsing, while SVM's due to their ability to handle high dimensional spaces were extensively used in sentiment analysis and text classification.

In the early 2000s, the use of Recurrent neural networks and Long Short-Term memory networks, led to a massive upward shift in the complexity and the coherence of the kinds of texts that could be generated. Particularly, LSTM models were able to learn long-term dependencies even in situations where the gap between useful information stretched over long sequences.

The current era is marked by the introduction of Transformers architecture in 2017. In the paper "Attention is all you need" (Vaswani et al.,2017), the concept of the transformer was unveiled, which was based on a self-attention mechanism that allowed models to focus on contextually relevant portions of the text to map complex and highly varied dependencies. This ability of self-transformers to capture dependencies and relationships between words, regardless of their distance within a sentence was a game changer in the field of NLP at the time. The paper marked a significant shift away from the use of RNNs and LSTM for tasks about sequence modeling. Also, unlike the previous generation of models like RNN and LSTM which could only process data sequentially, transformers were able to process all parts of a sentence in one go and could thereby support parallel processing which made it practically possible to train models on extremely large bodies of texts with lower training time.

Its successors namely, GPT (Generative Pre-Trained Transformer) by OpenAI and BERT (Devlin et al.,2019) (Bidirectional Encoder Representations from Transformers) by Google are what have led to the unleashing of this current era of generative AI in the field of NLP.

GPT 1 by OpenAI in 2018, was one of the first models to leverage transformers on language tasks and had 117 million parameters. It was trained on a dataset called the Book Corpus which consisted of around 7000 fiction books. The dataset was chosen because of its long paragraphs, to train the model to learn complex and longer-range features. The key innovation in this model was the use of unsupervised learning over a large corpus to create a general model (Radford et al.,2018). This general model was then fine-tuned across small task-specific datasets and was able to produce very good results across several NLP tasks like text summarization, question answering, and translation to name a few. This marked a distinct shift in the field of NLP away from NLP models that performed very specific tasks to the creation of a general model, which would then be finetuned according to the task to be performed.

GPT-2, was released fully in November 2019 and was a huge 1.5 billion parameter model that was trained on 8 million web pages (Radford et al.,2019). The corpus on which it was trained consisted of web pages linked to Reddit posts that had an upvote count of at least 3, which was then further refined by removing duplicate pages and links from Wikipedia. It is the training over this diverse dataset that allowed it to generate text that was coherent and relevant to the context at hand across a very broad set of topics. The capabilities of GPT 2 and the potential for its misuse were such that it sparked an outcry about the ethical considerations within the AI community. However, it had two major limitations i.e. it was unable to generate coherent and contextually relevant text of larger length and highly resource intensive in terms of the compute resources required. The ability of the model over longer text lengths was a significant issue, as the models would have repetitions, deviation from the context, and even contradictions for the text generated earlier.

GPT-3, was released by OpenAI in June 2020 and was an even bigger model than GPT 2 with 175 billion parameters. This was a truly large language model and its training corpus consisted of a filtered version of the common crawl dataset, Webtext 2 corpus, Wikipedia, and certain corpora of books. As a result of being trained over such a large dataset, it was able to perform zero-shot learning and few-shot learning (Brown et al.,2020). It was also able to code in languages like CSS, jax, python, and other languages.

Building upon this legacy, GPT-4 has pushed the boundaries further, with its increased parameter count and refined algorithms enabling even more nuanced and accurate text generation, closely mimicking human-like articulation and reasoning. Meanwhile, models like Gemini have harnessed bespoke architectures to cater to specialized tasks, optimizing for more efficient processing without compromising the depth of language comprehension. LLAMA, with its unique approach to language model training, showcases versatility in handling multiple languages and dialects, ensuring inclusivity in digital communication. These newer LLMs have expanded the scope of text generation to encompass more complex tasks such as summarizing intricate documents, generating code, and even creating educational content. Their relevance in today's data-driven world is underscored by their application in automating customer service, content creation, and assisting in decision-making processes (Davenport et al.,2022).

This research aims to undertake a comprehensive evaluation and comparative analysis of various prominent decoding methodologies utilized in text generation with a pre-trained GPT-2 model. By conducting a thorough examination of the inherent strengths and weaknesses inherent in these methods, the study endeavors to establish a set of metrics that will enable the identification of the most efficacious decoding technique. This new scheme will be designed to double as a tool for adversarial attacks, capable of effectively challenging and probing the robustness of text classification models.

The main contributions of the current work are fourfold. First, it introduces several decoding strategies for text generation including greedy search, beam search, Top-K sampling, Top-P sampling (nucleus sampling), contrastive search, and locally typical sampling, and evaluates its efficacy and effectiveness. Second, the performance of these methods is assessed in terms of text coherence, relevance, and diversity. Third, we provide the strengths and limitations of each decoding method discussed. Finally, we introduce a novel text generation scheme by modifying some of the existing approaches used in adversarial attacks on text classification models.

The remaining sections of the report are organized as follows: Section 2 presents some related works on text generation methods. Some useful and important terms relevant to text generation and language models are defined in Section 3. In Section 4, under methodology, theoretical background information on decoding methods for text generation, including greedy search, beam search, top-K sampling, and top-P sampling, is discussed. Section 5 presents detailed performance results of these decoding methods, along with an analysis of their efficacy and effectiveness. Section 6 concludes the report by summarizing key findings and proposing avenues for future research in the field of text generation decoding methods.

## 2. Related Work

The rise of Large Language Models (LLMs) and text generators marks a significant advancement in the domain of NLP. Rooted in deep learning architectures, these models demonstrate impressive capabilities in generating human-like text, grasping context, and performing various language-related tasks. This review aims to explore the progress, applications, challenges, and ethical considerations surrounding LLMs and text generators.

Over the past decade, large language models have undergone substantial development, driven by advancements in deep learning, computational resources, and data availability. Early models like Recurrent Neural Networks (RNNs) and Long Short-Term Memory (LSTM) networks laid the groundwork for understanding sequential data, particularly text. However, the introduction of transformer architectures by Vaswani et al. (2017), marked a breakthrough in NLP. Transformers revolutionized the field by enabling parallelization and efficiently capturing long-range dependencies, paving the way for larger and more powerful language models.

Subsequent models, built upon the Transformer architecture, such as the Generative Pre-trained Transformer (GPT) of OpenAI and Bidirectional Encoder Representations from Transformers (BER) of Google, further expanded the capabilities of LLMs. These models were trained on an extremely large volume of text corpora and optimized for specific tasks, achieving state-of-the-art performance in areas like language translation and sentiment analysis.

In the early stages of NLP development, researchers relied on benchmarking tests for evaluating the performance of language models. These tests focused on vocabulary and grammar, including parsing for syntactic conformance and disambiguation of the senses of words. The introduction of the MUC evaluation in the early 1990s was a significant event (Grishman & Sundheim, 1996) that effectively propelled advancement in information extraction techniques. Deep learning models led to broader benchmarks such as SNLI (Bowman et al., 2015) and SQuAD (Rajpurkar et al., 2016) were introduced. These frameworks were able to evaluate system performance and also provided copious training data. Moreover, these benchmarks assigned scores based on selected metrics that helped in evaluating the accuracies of specific tasks.

With the invention of large language models that are pre-trained such as BERT (Devlin et al., 2019), the techniques for evaluating these new generalized models have gradually adapted. In response, the NLP community has organized numerous shared tasks and challenges, including SemEval (Nakov et al., 2019), CoNLL (Tjong et al., 2003), SuperGLUE (Wang et al., 2019), and XNLI (Conneau et al., 2018). These initiatives involve evaluating the overall performance of a framework by aggregating the performance scores of each model. As a result, there has been a continued effort in refining the methodologies of evaluating NLP frameworks, and designing a holistic and dynamic environment for researchers so that they can evaluate the efficacies of different systems.

As LLMs continue to expand in size, they have demonstrated impressive performance, challenging optimized pre-trained models. This has led to a shift in the landscape of evaluation, moving away from general benchmarks set for specific tasks towards assessments focused on capabilities. The distinctions between different downstream tasks are becoming less clear, and there has been an expansion in the variety of benchmarks of evaluation constructed to assess reasoning, knowledge, and several other abilities. These benchmarks do not rely on training data and aim to provide a holistic framework for evaluating the capability of a model in different scenarios (Hendrycks et al., 2021; Zhong et al., 2023; Zhang et al., 2023; Li et al., 2023).

The swift uptake of LLMs by users and researchers is evident in the remarkable success of ChatGPT (OpenAI, 2022), which had more than 100 million users within a couple of months of its release. This extraordinary rise in the user base highlights the potential of these models, spanning various applications such as text generation (Brown et al., 2020), generation of program codes (Chen et al., 2021), and utilization of tools (Nakano et al., 2021). However, despite their promising capabilities, these models have serious issues regarding the risks associated with deploying them extensively without extensive tests. Adverse issues like model biases, misinformation dissemination, and privacy breaches, demand careful consideration. In response, there has been a distinct focus in research on empirically assessing how the alignment of LLMs with the preferences of humans.

PLMs are neural networks that undergo training on extensively large datasets without specific labeling, enabling them to later be adapted for specific tasks. Research indicates that PLMs are capable of capturing vast volumes of linguistic knowledge based on their designs and parameters. This may lead to improvements in language comprehension and the quality of generated text.

Given the remarkable success of the Transformer architecture, nearly all PLMs are built upon it. For instance, prominent PLMs like GPT and BERT initially adopt the decoder and encoder architectures of Transformer, respectively. Various PLMs such as RoBERTa, XLNET, and BAERT vary in design, with some variants derived from BERT, while others are PLMs of encoder-decoder types. Recent research suggests that enhancing model parameters can significantly improve PLM performance, leading to the design of massively large-scale PLMs like GPT-3 (175B), PANGU (200B), GShard (600B), and Switch-Transformers (1.6T). Additionally, language models are tailored for various tasks like named entity recognition.

Pre-existing PLMs utilized for text generation typically employ a single Transformer or a Transformer-based encoder-decoder structure as their core framework. For example, GPT-3 (Brown et al., 2020) and UniLM (Dong et al., 2019) utilize an encoder or decoder on the Transformer to handle both input encoding and output decoding concurrently. These models consist of three main variants: masked language models, causal LMs, and prefix LMs, each employing distinct strategies. A detailed overview of these four variants is provided below.

*Masked Language Models* (MLMs) employ a full-attention Transformer encoder, enabling them to undergo pre-training using the masked language modeling (MLM) task, which involves bidirectional prediction of masked tokens. BERT (Devlin et al., 2019) is a prominent example of such a model, widely utilized in natural language understanding (NLU). However, the discrepancy between the pre-training task of masked LMs and the subsequent generation function limits their application in text generation tasks (Yang et al., 2019). Instead, masked LMs are commonly utilized as the encoder component for text generation, leveraging their robust bidirectional encoding capabilities. For instance, Rothe et al. (2020) proposed initializing an integrated encoder-decoder with BERT (Devlin et al., 2019), achieving performance at par with text-generation PLMs.

*Causal Language Models* (LMs) adopt a diagonal mask matrix, akin to the Transformer decoder. These models are used in modeling language, aiming to identify the likelihood of a series of words in a sequence occurring in a sentence. In text generation, causal LMs excel in selecting the next word based on preceding words. The language model GPT (Radford et al., 2018) pioneered the use of causal LMs. Subsequent advancements, including GPT-2 (Radford et al., 2019), investigated the possibility of transfer capabilities of language models in text generation, emphasizing the importance of the availability of abundant data. Additionally, GPT-3 (Brown et al., 2020) showcased the significant impact of scaling up model parameters on enhancing downstream generation tasks, often requiring only a few prompts. Conversely, CTRL (Keskar et al., 2019) introduced a conditional causal LM tailored for text generation. Despite their suitability for the task, causal LMs have several constraints. These models strictly encode tokens from left to right, overlooking bidirectional input information. Moreover, they are not optimized for sequence-to-sequence generation tasks, resulting in suboptimal results in text summarization and language translation (Radford et al., 2019).

*Prefix Language Models* are based on architecture using a single Transformer, utilizing bidirectional encoding for input and a left-to-right generation method for output. By employing a mask with mixed attention, tokens within the input text $x$ can interact with one another, while tokens in the target text $y$ are restricted to attending solely to tokens at the input and previously used tokens. UniLM (Dong et al., 2019) is notably the pioneering LM, that prefixes an attention mask for conditional generation tasks. Further advancements, such as UniLMv2 (Bao et al., 2020) and GLM (Du et al., 2021), refine the basic strategy of masking by introducing permutation-based modeling from XLNet (Yang et al., 2019). Despite their numerous advantages, a comparison conducted by Raffel et al. (2020) suggests that explicitly integrating encoder-decoder attention in Transformer-based encoder-decoder LMs is more effective in capturing conditional dependencies compared to single-transformer prefix LMs.

*Encoder-decoder language models* follow the conventional Transformer architecture for generating text, consisting of stacks of encoder and decoder layers. During the pre-training stage, methods like MASS (Song et al., 2019) and ProphetNet (Qi et al., 2020) utilized a sequence with a single masked segment as input for the encoder. Subsequently, the decoder generated the masked tokens in a sequential, auto-regressive manner. T5 (Raffel et al., 2020), on the other hand, employed a distinct strategy by randomly replacing several spans within the source text with various special tokens. The decoder then predicted each replaced span in sequence. Meanwhile, BART (Lewis et al., 2020) underwent pre-training using a denoising auto-encoder (DAE) technique. Here, the model learned to reconstruct the original text from altered versions induced by different noise methods, including sentence permutation and token deletion.

When adjusting PLMs in varied text generation tasks, it's crucial to factor in particular language characteristics. In the following, three primary properties that are frequently sought in text generation are discussed.

*Relevance:* As outlined in linguistic literature (Li et al., 2021), pertains to how closely the thematic meaning conveyed in generated text aligns with the input text. In dialogue systems, for example, the generated responses must be relevant to the preceding utterances and other contextual elements such as speaker traits and discourse topics. Unlike traditional neural generative models, pre-trained language models (PLMs) employ a more robust multi-layer cross-attention mechanism to capture semantic connections between input and output, thus enhancing the relevance of the generated text to the input data. An excellent illustration of this is seen in DialoGPT (Zhang et al., 2020), an extension of the auto-regressive language model GPT-2. DialoGPT is trained on extensive dialogue pairs/sessions, allowing it to understand the joint distribution of conversation history and response for generating contextually relevant replies. Additionally, Zeng & Nie (2020) utilized a masked language modeling objective to generate responses based on various dialogue contexts. They introduced a TF-IDF-based masking approach to selectively mask tokens that are more relevant to the context, enabling PLMs to produce contextually relevant expressions rather than generic language patterns. Moreover, they proposed a non-parametric attention-based gating mechanism to dynamically switch between generating general words or contextually relevant words at each position.

*Faithfulness* is a crucial element in language when it comes to generating text, indicating the importance of ensuring that generated content accurately reflects the meaning of the input. In applications such as text summarization, the aim is to produce text that effectively captures the main intent of the input text. This is known as the faithfulness of text. This concept also extends to ensuring that the generated text remains consistent with factual knowledge about the world. Achieving faithfulness in text generation requires PLMs to accurately comprehend the underlying semantics of the input and have access to adequate world knowledge to address the task at hand.

Research has demonstrated the impressive natural language understanding capabilities of PLMs in extracting core semantics from text (Devlin et al., 2019), as well as their ability to encode extensive world knowledge (Jiang et al., 2020). This knowledge can be utilized to enhance faithfulness by integrating knowledge into the text generated by the LM. For instance, Kryscinski et al. (2018) employed a network of context within the decoder of the PLM in extracting the most relevant segments from the source document, thereby improving the faithfulness of the generated summaries. Furthermore, several studies have proposed incorporating additional loss functions alongside text generation loss to encourage faithfulness (Rothe et al., 2020; Yang et al., 2020a). For example, Yang et al. (2020a) fine-tuned PLMs using a theme modeling loss to ensure that the generated summaries maintain semantic alignment with the original article for faithful generation.

*Order-Preservation:* Consistent arrangement of semantic elements, such as words or phrases, between input and output texts, is a defining characteristic in Natural Language Processing (NLP) termed Order-Preservation. This feature is relevant to various text generation tasks, including paraphrasing and machine translation. Maintaining the sequence of phrases between the source and target texts is essential for accurate machine translation tasks. Methods that involve the alignment of words have been widely used in this field to maintain this property. Notably, the Code-Switching Pre-training (CSP) technique (Yang et al., 2020b) stands out. CSP starts by extracting alignment details between word pairs from monolingual datasets containing the source and target languages. To meet the requirements of order preservation in machine translation, CSP iteratively improves PLMs by making robust predictions of fragments of sentences in the source language based on the corresponding fragments of sentences in the target language. Additionally, addressing the limitation of the approach to the alignment of discrete words, a method focusing on the alignment of words in a continuous stream has been proposed to bolster the preservation of words. Wada & Iwata (2018) concentrate on word alignment representations across languages by mapping the embedding of words to a shared latent space. A novel scheme named mRASP is proposed by Lin et al. (2020) to ensure the alignment of words and phrases with akin meanings over several languages. The objective of the project is to build an efficient and accurate text generator model that will be precisely able to understand the impact of various tuning parameters of the GPT language model, and generate text sequences of arbitrarily large length with a high level of relevance based on a prompt given by the user. Several versions of the models will be designed having different sets of parameters with varying levels of precision in the generated texts and their relevance.

## 3. Definitions of Some Relevant Terms

In this Section, some important and relevant terms in the field of text generation are provided for the benefit of the readers.

- prompt: The initial text or context provided to a language model to guide its generation process.
- num_beams: The number of beams (alternative sequences) considered during beam search decoding. A higher value increases diversity but also computation time.
- max_length: The maximum length (in tokens) for the generated output. It restricts the length of the generated text.
- temperature: Controls the diversity of generated text.
- top-k: Identifies the k tokens that have the highest probability of being generated in the current position.
- top-p: Sets the cumulative probability threshold for token sampling.
- input_ids: Input sequence encoded as token IDs
- attention_mask: Masks padding tokens to focus on relevant input
- pad_token_id: Specifies padding token ID for generated sequences
- no_repeat_ngram_size: Prevents repetition of n-grams of specified size
- do_sample: Enables token sampling during generation

## 4. Methodology

In this Section, theoretical background information on text generation methods, including their working principles, strengths, and weaknesses are presented. The text-generation methods considered are as follows: (i) Greedy Search, (ii) Beam Search, (iii) Top-K Sampling, (iv) Top-P Sampling, (v) Contrastive Search, and (iv) Locally Typical Sampling.

## 4.1 Greedy Search

Greedy Search is one of the simplest decoding methods used in text generation tasks with neural language models like GPT-2. The principle behind greedy search is straightforward: at each step of generating text, the model selects the token with the highest probability as the next token. This process continues iteratively until a predefined stopping criterion is met, such as reaching a maximum length or encountering an end-of-sentence token. The primary outcome of Greedy Search is a generated sequence of tokens that maximizes the likelihood of the next token at each step. This method tends to produce outputs relatively quickly since it always selects the most probable token.

Mathematically, greedy search involves calculating the conditional probability distribution over the vocabulary given the previously generated tokens. Let's denote the generated tokens so far as $w_1$, $w_2, \dots$ $w_{t-1}$, and we aim to predict the next token $w_t$. The model computes the probability $P(w_t|w_{t-1}, w_{t-2}, \dots \dots w_1)$ for each token in the vocabulary and selects the token with the highest probability as the next one. The method is represented in (1).

$$W_t = argmax\ wP(w_t|w_{t-1}, w_{t-2}, \dots \dots w_1) \quad (1)$$

Let us consider that we want to generate text based on the prompt *"The cat sat on the."* Let us assume the vocabulary consists of three tokens: "mat," "chair," and "rug," with the associated probabilities as follows:

P("mat"|"The cat sat on the")=0.6

P("chair"|"The cat sat on the")=0.3

P("rug"|"The cat sat on the")=0.1

In this case, greedy search would select "mat" as the next token because it has the highest probability. While greedy search selects the locally most probable token at each step, it doesn't consider the global context or future tokens, potentially leading to suboptimal sequences. In this case, the model may miss out on more contextually appropriate choices in favor of immediate high-probability tokens. This can be explained by an example in Figure 1.

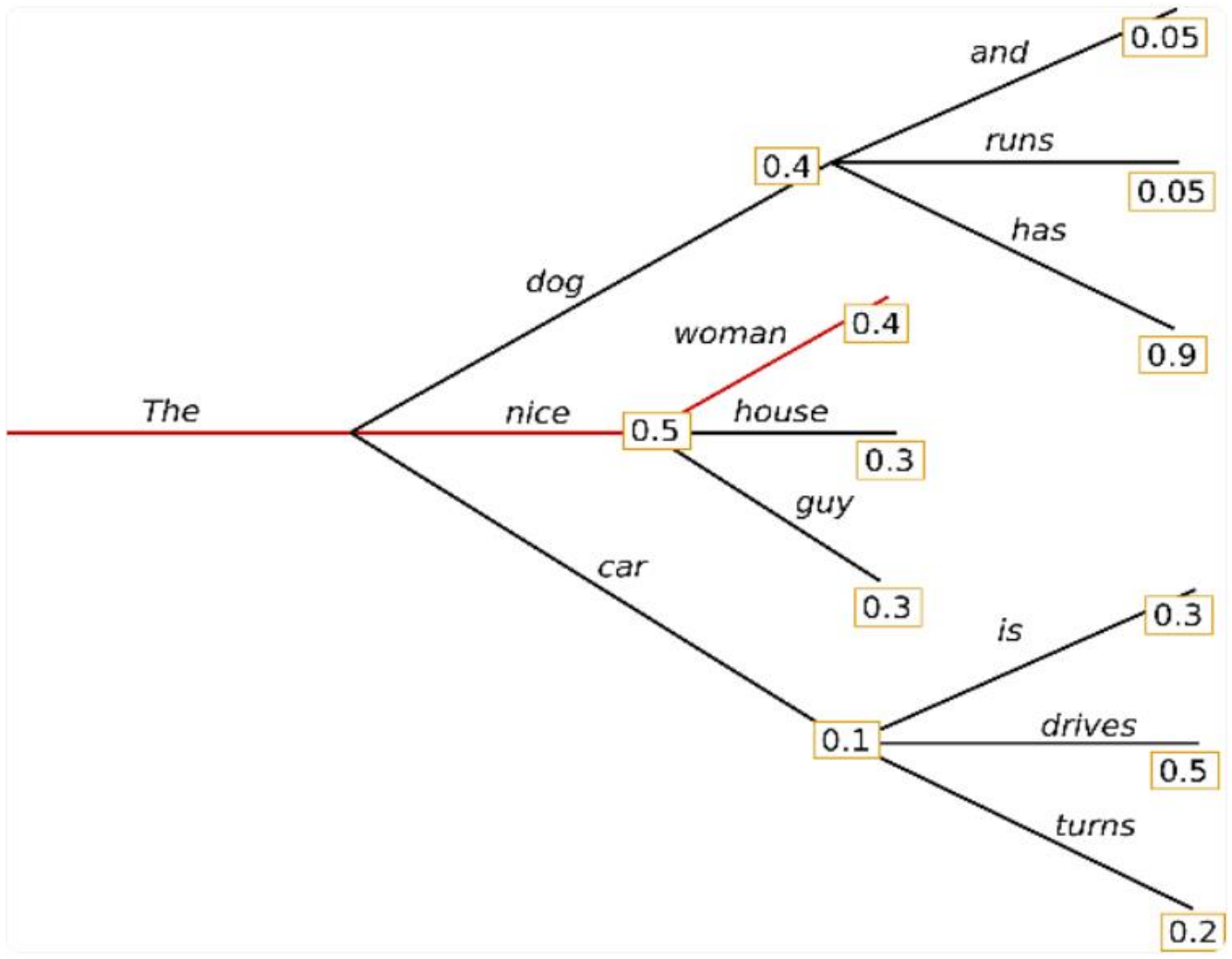

Figure 1: Illustration of the working principle of Greedy Search

Greedy search, while efficient and straightforward, exhibits significant limitations that affect the quality and diversity of generated text.

- One prominent drawback is its tendency to produce repetitive or generic outputs due to its reliance on selecting the most probable token at each step.
- By favoring local probabilities, greedy search often overlooks less probable yet potentially more diverse options, resulting in monotonous and predictable sequences.

- Moreover, its insensitivity to global context. Greedy search fails to capture long-term dependencies in the text, focusing solely on immediate surroundings without considering the overarching narrative or thematic coherence. Consequently, this narrow scope may lead to suboptimal sequences that lack contextual relevance and fail to capture the nuanced nuances of language.

This limitation highlights the importance of considering alternative decoding methods like beam search or sampling techniques to generate more diverse and contextually relevant text.

## 4.2 Beam Search

Introduced by Fred Jelinek (1969), beam search is a more sophisticated decoding method compared to greedy search, aiming to address its limitations by exploring multiple potential continuations at each step. Instead of selecting only the single most probable token, beam search maintains a set of candidate sequences, called the beam and expands them simultaneously. This exploration process continues iteratively until a predefined stopping criterion is met. Beam search aims to strike a balance between coherence and diversity in the generated text. Exploring multiple hypotheses simultaneously, it tends to produce more diverse outputs compared to Greedy Search while still maintaining some level of coherence. The primary outcome of beam search is a set of candidate sequences, each representing a potential continuation of the input text.

At each step of beam search, the model generates a set of candidate sequences based on the tokens selected in the previous step. These candidate sequences are ranked according to their cumulative probabilities. The top-*N* sequences with the highest cumulative probabilities, where *N* is the beam width, are retained, and the process continues iteratively.

Let $B_t$ represent the set of candidate sequences at step *t*, where each candidate sequence *b* consists of a sequence of tokens $b = (b_1, b_2, \dots\dots b_t)$.

Given the set of candidate sequences $B_t$ at step *t*, the model calculates the conditional probability of extending each candidate sequence with each token in the vocabulary. Let $P(b|b_1, b_2, \dots . b_{t-1}, w)$ denote the conditional probability of extending candidate sequence *b* with token *w* at step *t*. The model then computes the cumulative probabilities of all possible extensions of each candidate sequence by summing the conditional probabilities of each token. Let $P_c(b)$ represent the cumulative probability of candidate sequence *b* at step *t* as represented in (2)

$$P_c(b) = P(b|b_1, b_2, \dots \dots b_{t-1}) = \sum wP(b|b_1, b_2, \dots\dots b_{t-1}, w) \quad (2)$$

Next, the top-$N$ candidate sequences with the highest cumulative probabilities are retained, where N is the beam width. These top-N candidate sequences form the set of candidate sequences for the next step, $B_{t+1}$.

The process continues iteratively until a predefined stopping criterion is met, such as reaching a maximum length or encountering an end-of-sequence token.

Let us consider the prompt "The cat sat on the." Using beam search with a beam width of 3 and a vocabulary consisting of "mat," "chair," and "rug" with associated probabilities, as follows:

P("mat"|"The cat sat on the.")=0.6

P("chair"|"The cat sat on the.")=0.3

P("rug"|"The cat sat on the.")=0.1

At the beginning of the process, the beam is empty.

Then the model generates the first set of candidate sequences by extending the empty beam with each token from the vocabulary. The top-3 candidate sequences with the highest cumulative probabilities are retained.

*Candidate sequences at step 1:*

Candidate 1: "mat" with probability Pc("mat")=0.6

Candidate 2: "chair" with probability Pc("chair")=0.3

Candidate 3: "rug" with probability Pc("rug")=0.1

The model generates the next set of candidate sequences by extending each retained candidate sequence from the previous step with each token from the vocabulary. The top-3 candidate sequences with the highest cumulative probabilities are retained.

*Candidate sequences at step 2* (assuming each previous candidate extends with the same probabilities):

Candidate 1: "mat" → "mat" with probability Pc("mat"→"mat")=0.6×0.6=0.36

Candidate 2: "mat" → "chair" with probability Pc("mat"→"chair")=0.6×0.3=0.18

Candidate 3: "mat" → "mat" with probability Pc("mat"→"rug")=0.6×0.1=0.06

The top-3 candidate sequences with the highest cumulative probabilities are retained for the next step. Final selected candidate sequences after step 2:

Candidate 1: "mat" → "mat"

Candidate 2: "mat" → "chair"

Candidate 3: "mat" → "rug"

Thus, beam search with a beam width of 3 explores three possible continuations of the input text by considering the probabilities of each token in the vocabulary. It selects the most probable sequences among them, allowing for a balance between coherence and diversity in generated text.

Thus beam search uses multiple candidate sequences simultaneously, allowing for a more comprehensive exploration of possible continuations and reducing the likelihood of unexpected or divergent outputs. Greedy search, on the other hand, tends to favor locally optimal choices without considering alternative paths, potentially resulting in more unpredictable outcomes.
Despite its improvements over greedy search, beam search has its following limitations.

- Fixed Beam Width: Beam search requires specifying the beam width, which influences the trade-off between diversity and coherence. Choosing an appropriate beam width can be challenging and may require tuning for different tasks and models.

- Risk of Local Optima: Early in the generation process, beam search may favor highly probable but suboptimal sequences, potentially missing out on more diverse or coherent alternatives.

## 4.3 Top-K Sampling

Introduced by Fan et al. (2018), Top-*K* sampling serves as a probabilistic approach in text generation tasks, offering a balance between randomness and control over the diversity of generated sequences. Unlike deterministic methods like greedy search, Top-*K* sampling introduces variability by selecting the next token from the Top-*K* most likely candidates at each step. This allows for the exploration of diverse continuations while still prioritizing tokens with higher probabilities.

Top-*K* sampling involves computing the conditional probability distribution over the vocabulary given the preceding tokens in the sequence. Let's denote the generated tokens so far as $w_1, w_2, \ldots . w_{t-1}$, and we aim to predict the next token $w_t$. The model calculates the probability $P(w_t | w_1, w_2, \ldots \ldots w_{t-1})$ for each token in the vocabulary and selects one token from the Top-*K* candidates based on their probabilities. This is represented in (3).

$$w_t \sim Categorical(P(w | w_1, w_2, \ldots \ldots w_{t-1}) \quad (3)$$

where $w \in Top - K$

This stochastic process introduces randomness into the generation process, allowing for the exploration of diverse sequences while maintaining a level of control over the likelihood of the selected tokens. This process is illustrated in Figure 2.

Consider generating text based on the prompt "The cat sat on the." Assume a vocabulary consisting of three tokens: "mat," "chair," and "rug," with associated probabilities as follows:

P("mat"|"The cat sat on the")=0.6

P("chair"|"The cat sat on the")=0.3

P("rug"|"The cat sat on the")=0.1

In this example, Top-K sampling would randomly select one token from the Top-K candidates ("mat" and "chair") based on their probabilities.

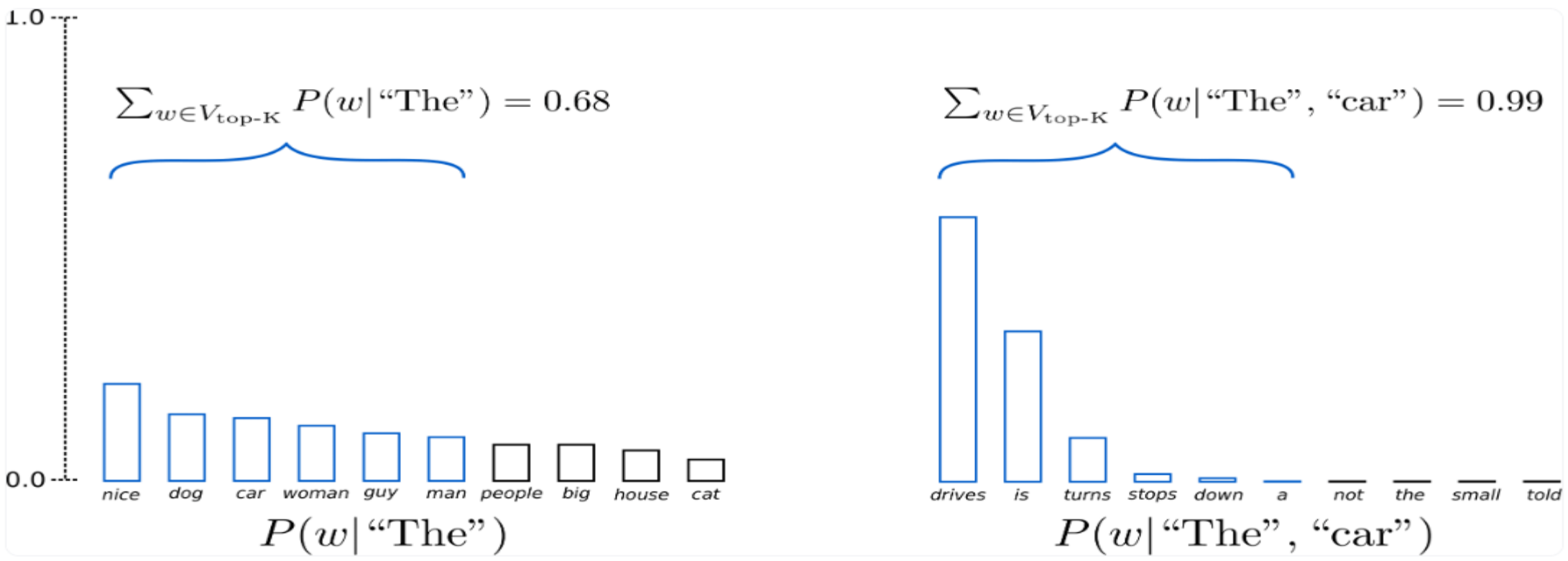

Figure 2: Illustration of the working principle of Top-K Sampling

Despite its advantages, top-K sampling has several limitations:

- Risk of Repetition: Top-K sampling may still produce repetitive sequences, especially when the top-K candidates contain similar tokens with high probabilities. This can lead to the generation of monotonous or redundant text, diminishing the quality of the generated output.

- Lack of Coherence: While top-K sampling promotes diversity, it may result in the generation of incoherent or disjointed sequences, particularly when the selected tokens do not align well with the context of the preceding sequence. This can undermine the overall fluency and coherence of the generated text.

- Hyperparameter Sensitivity: The performance of top-K sampling is sensitive to the choice of the hyperparameter K, which determines the size of the candidate pool. Selecting an inappropriate value for K can impact the trade-off between diversity and quality, potentially leading to suboptimal results.

## 4.4 Top-P Sampling

Proposed by Holtzman et al. (2019), Top-$P$ sampling, also known as nucleus sampling or probabilistic sampling, is a probabilistic approach in text generation tasks, providing a balance between diversity and control over the likelihood of generated sequences. Similar to Top-$K$ sampling, it introduces variability by selecting the next token from the Top-$P$ most likely candidates at each step. This method dynamically adjusts the size of the candidate pool based on a cumulative probability threshold, allowing for the exploration of diverse continuations while ensuring that the selected tokens collectively account for a significant portion of the probability mass.

Top-$P$ sampling involves computing the conditional probability distribution over the vocabulary given the preceding tokens in the sequence. Let's denote the generated tokens so far as $w_1, w_2, \dots. w_{t-1}$, and we aim to predict the next token $w_t$. The model calculates the probability $P(w_t|w_1, w_2, \dots.. w_{t-1})$ for each token in the vocabulary and selects one token from the Top-$P$ candidates based on their probabilities. This is represented in (4)

$$w_t \sim Categorical(P(w|w_1, w_2, \dots.. w_{t-1}) \quad (4)$$

where $w \in Top - P$

This stochastic process introduces randomness into the generation process, allowing for the exploration of diverse sequences while maintaining a level of control over the likelihood of the selected tokens.

Let us consider generating text based on the prompt “The cat sat on the.” Assume a vocabulary consisting of three tokens: “mat,” “chair,” and “rug,” with associated probabilities as follows:

P("mat"|"The cat sat on the.")=0.6

P("chair"|"The cat sat on the.")=0.3

P("rug"|"The cat sat on the.")=0.1

In this example, Top-*P* sampling would dynamically adjust the candidate pool based on a cumulative probability threshold. Let's assume the cumulative probability threshold is set to 0.8. The model would select tokens "mat" and "chair" since their cumulative probabilities sum up to more than 0.8. The approach is pictorially depicted in Figure 3.

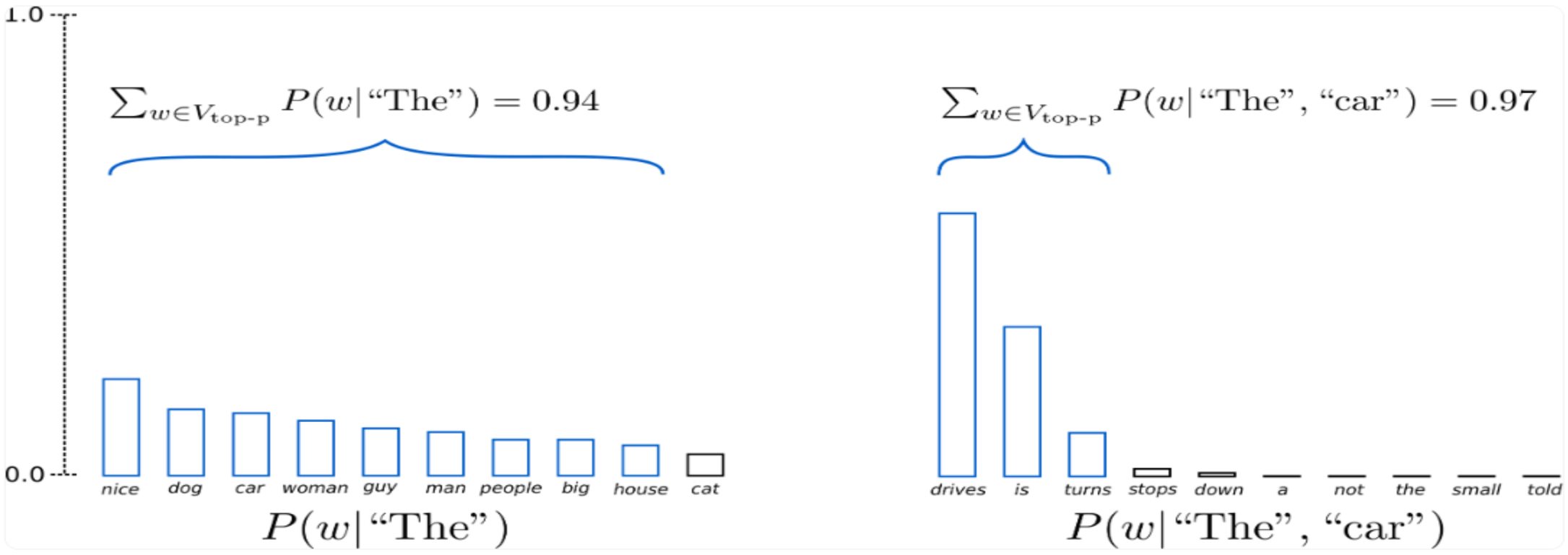

Figure 3: Illustration of the working principle of Top-P Sampling

Despite its advantages, top-P sampling has several limitations:

- Risk of Repetition: Similar to top-K sampling, top-P sampling may still produce repetitive sequences, especially when the top-P candidates contain similar tokens with high probabilities. This can lead to the generation of monotonous or redundant text, diminishing the quality of the generated output.

- Lack of Coherence: While promoting diversity, top-P sampling may result in the generation of incoherent or disjointed sequences, particularly when the selected tokens do not align well with the context of the preceding sequence. This can undermine the overall fluency and coherence of the generated text.

- Hyperparameter Sensitivity: The performance of top-P sampling is sensitive to the choice of the cumulative probability threshold P, which determines the size of the candidate pool. Selecting an inappropriate value for P can impact the trade-off between diversity and quality, potentially leading to suboptimal results.

## 4.6 Contrastive Search

Contrastive search operates by selecting the most suitable token from a candidate set based on the contrastive loss function. It aims to mitigate model degeneration by encouraging diverse and semantically coherent text generation. The key idea is to incorporate a contrastive objective during decoding to penalize repetitive and semantically inconsistent outputs.

The objective function of contrastive search in text generation aims to optimize the similarity between the generated text and a reference text while minimizing the similarity between the generated text and negative examples. The contrastive objective function typically consists of two components: a similarity term and a dissimilarity term. The general formulation of the objective function for contrastive search is given by (5).

$$OBJ(CS) = \alpha * Similarity\ (G, R) + \beta * Dissimilarity(G, N) \tag{5}$$

Where $G$ represents the generated text, $R$ is the reference text, $N$ stands for a set of negative examples, Similarity ($G$, $R$) measures the similarity between the generated text $G$ and the reference text $R$, Dissimilarity ($G$, $N$) measures the dissimilarity between the generated text $G$ and the negative examples $N$., $\alpha$ and $\beta$ are the hyperparameters that control the trade-off between maximizing similarity and minimizing dissimilarity.

The objective function encourages the model to produce text that is semantically similar to the reference text while being dissimilar to negative examples. This encourages the generated text to capture the same semantic content and style as the reference text while avoiding undesirable characteristics present in the negative examples. The choice of similarity and dissimilarity metrics may vary depending on the specific task and domain. Common similarity metrics include cosine similarity, BLEU score, BERTScore, or other measures of semantic similarity between text sequences. Dissimilarity metrics may include measures of edit distance, semantic contrast, or dissimilarity scores computed for negative examples.

Despite its effectiveness in mitigating model degeneration, contrastive search may encounter limitations in scenarios where candidate tokens are limited.

- The contrastive loss function may fail to adequately capture semantic coherence in certain contexts, leading to suboptimal token selection.
- The computational overhead of calculating contrastive scores for a large candidate set may impact decoding efficiency, particularly in real-time applications.

## 4.7 Locally Typical Sampling

Proposed by Meister et al. (2023), Locally typical sampling is a novel decoding strategy designed to address the limitations of traditional sampling methods in text generation tasks. It aims to generate text that closely matches the information content expected by humans given the prior context. It operates by constraining the sampling distribution to words whose negative log-probability falls within a certain absolute range from the conditional entropy of the model at each time step. This range is determined by a hyperparameter τ, representing the amount of probability mass from the original distribution that is considered. The algorithm dynamically adjusts τ per word, making it less sensitive to hyperparameter choice.

The formulation of locally typical sampling involves optimizing a subset of the vocabulary set $C_{y<t}$ to minimize the conditional entropy $H(Y_t|\boldsymbol{Y}_{<t} = y_{<t})$ while ensuring that the probability of sampled words exceeds a threshold τ. Mathematically, this process is represented in (6).

$$minimize \ \sum_{y \in C'(y_{<t})} |H(Y_t \,|\boldsymbol{Y}_{<t} = \boldsymbol{y}_{<t}) + log\ p(y|\boldsymbol{y}_{<t})| \quad (6)$$

$$\text{where } C(y_{<t}) \in P(\overline{v})$$

$$\text{Subject to: } \sum_{y \in C(y_{<t})} p(y|\boldsymbol{y}_{<t}) \geq \tau$$

where, $C(y_{<t})$ is the truncation set representing the solution to the optimization problem

$C(y_{<t}) \in P(\overline{v})$ is a context-dependent subset of the vocabulary

$H(Y_t|Y_{<t} = y_{<t}$ is the conditional entropy of the model at step *t* given the prior context

$p(y|y_{<t})$ is the probability of word $y$ given the prior context

$\tau$ is the hyperparameter controlling the amount of probability mass considered from the original distribution.

While locally typical sampling offers a promising approach to generating text that aligns closely with human expectations, it also has the following limitations.

- Computational Complexity: Although locally typical sampling can be implemented with similar efficiency as other sampling methods such as the nucleus or top-k sampling, it still involves additional computational overhead due to the optimization process and dynamic adjustment of the hyperparameter $\tau$.

- Hyperparameter Sensitivity: The effectiveness of locally typical sampling may depend on the choice of the hyperparameter $\tau$, which governs the range of acceptable probabilities for sampled words. Tuning this hyperparameter optimally for different tasks and datasets may require careful experimentation.

# 5. Results and Analysis

In this Section, we present detailed results of the text-generation methods and compare their performances on several standard metrics of text generation. In Section 5.1, we present several screenshots of code snippets of the programs for the search methods and outputs produced by them. In Section 5.2, the performance results of the text-generation methods are compared, and detailed results are presented.

## 5.1 Performance Results of the Text-Generators

In this Section, the code snippets and the output in the form of text generated by the text-generators are presented. For each text-generator, the code snippet of the program is first presented followed by the specimen output text yielded by it.

```python
def greedy_search(prompt, max_length=200):
    input_ids = tokenizer.encode(prompt, return_tensors="pt").to(device)
    attention_mask = torch.ones_like(input_ids)

    output = model.generate(
        input_ids=input_ids,
        attention_mask=attention_mask,
        pad_token_id=tokenizer.eos_token_id,
        max_length=max_length,
        do_sample=False
    )
    return tokenizer.decode(output[0], skip_special_tokens=True)
```

Figure 4: The program code snippet for Greedy Search

```python
prompt = "Can you explain the concept of artificial intelligence?"
print("Greedy Search (Vanilla):", greedy_search(prompt))
```

```
Greedy Search (Vanilla): Can you explain the concept of artificial intelligence?

I think it's a very interesting concept. I think it's a very interesting concept. I think it's a very
interesting concept. I think it's a very interesting concept. I think it's a very interesting concept. I
think it's a very interesting concept. I think it's a very interesting concept. I think it's a very
interesting concept. I think it's a very interesting concept. I think it's a very interesting concept. I
think it's a very interesting concept. I think it's a very interesting concept. I think it's a very
interesting concept. I think it's a very interesting concept. I think it's a very interesting concept. I
think it's a very interesting concept. I think it's a very interesting concept. I think it's a very
interesting concept. I think it's a very interesting concept. I think it's a very interesting concept. I
think it's a very interesting concept.
```

Figure 5: The specimen output text generated by Greedy Search

```python
def beam_search(prompt, num_beams=5, max_length=200):
    input_ids = tokenizer.encode(prompt, return_tensors="pt").to(device)
    attention_mask = torch.ones_like(input_ids)
    output = model.generate(input_ids, attention_mask=attention_mask,
                            pad_token_id=tokenizer.eos_token_id,max_length=max_length,
                            num_return_sequences=1, num_beams=num_beams)
    return tokenizer.decode(output[0], skip_special_tokens=True)
```

Figure 6: The program code snippet for Beam Search

```
prompt = "Can you explain the concept of artificial intelligence?"
print("Beam Search (Vanilla):", beam_search(prompt))


Beam Search (Vanilla): Can you explain the concept of artificial intelligence?

I think it's a very interesting concept. I think it's a very interesting concept. I think it's a very
interesting concept. I think it's a very interesting concept. I think it's a very interesting concept. I
think it's a very interesting concept. I think it's a very interesting concept. I think it's a very
interesting concept. I think it's a very interesting concept. I think it's a very interesting concept. I
think it's a very interesting concept. I think it's a very interesting concept. I think it's a very
interesting concept. I think it's a very interesting concept. I think it's a very interesting concept. I
think it's a very interesting concept. I think it's a very interesting concept. I think it's a very
interesting concept. I think it's a very interesting concept. I think it's a very interesting concept. I
think it's a very interesting concept.
```

Figure 7: The specimen output text generated by Beam Search

```
def top_k_sampling_search(prompt, temperature=1.0, top_k=50, max_length=200):
    input_ids = tokenizer.encode(prompt, return_tensors="pt").to(device)
    attention_mask = torch.ones_like(input_ids)
    output = model.generate(input_ids, attention_mask=attention_mask,
                            pad_token_id=tokenizer.eos_token_id, max_length=max_length,
                            num_return_sequences=1, do_sample=True, temperature=temperature,
                            top_k=top_k)
    return tokenizer.decode(output[0], skip_special_tokens=True)
```

Figure 8: The program code snippet for Top-K Sampling

```
prompt = "Can you explain the concept of artificial intelligence?"
print("Top K sampling (Vanilla):", top_k_sampling_pipeline(prompt))


Top K sampling (Vanilla): Can you explain the concept of artificial intelligence?

I think there were an increasing amount of philosophers who were pushing back on computer programming in
much the same way that they have about human intelligence. One of the things that's very powerful about
human brains is that they're highly intelligent.

What's also a very powerful thing about computers is that they allow to create things from other
materials. So we can have computers that can think for hundreds of thousands of years, that have been
built by human beings. You can get there by going back in time and reading books by Alan Turing about how
the computer was designed to see different things at different times. We're in this time of artificial
intelligence.

The thing about human intelligence is that if we get a human going forward in all sorts of ways, all they
can do is look at the future and see if they find something different about themselves in it.
```

Figure 9: The specimen output text generated by Top-K Sampling

```python
def top_p_sampling_search(prompt, temperature=1.0, top_p=0.9, max_length=200):
    input_ids = tokenizer.encode(prompt, return_tensors="pt").to(device)
    attention_mask = torch.ones_like(input_ids)
    output = model.generate(input_ids, attention_mask=attention_mask,
                            pad_token_id=tokenizer.eos_token_id, max_length=max_length,
                            num_return_sequences=1, do_sample=True, temperature=temperature,
                            top_p=top_p)
    return tokenizer.decode(output[0], skip_special_tokens=True)
```

Figure 10: The program code snippet for Top-P Sampling

```
prompt = "Can you explain the concept of artificial intelligence?"
print("Top p sampling (Vanilla):", top_p_sampling_search(prompt))

The concept of artificial intelligence is a new kind of thinking. You can think about it like the idea of
an algorithm that is going to be able to solve any problem, but it's going to have to learn something.
That is, the human brain is going to be able to take care of the human body and keep its health. That's a
huge part of what we are talking about here. And, you know, what we have to do is, you know, you can't
just have the brain of an artificial intelligence. It's going to have to learn a whole bunch of things,
and it's going to have to learn some basic things, but you know, we have to be able to use those, and
that is going to be very interesting.
What do you think of the rise of AI in the next few years? What do you think about that? The rise of AI
in the next few years.
```

Figure 11: The specimen output text generated by Top-P Sampling

```python
def contrastive_search(prompt, penalty_alpha=0.9, max_length=200):
    input_ids = tokenizer.encode(prompt, return_tensors="pt").to(device)
    attention_mask = torch.ones_like(input_ids)
    pad_token_id = tokenizer.eos_token_id
    output = model.generate(input_ids, attention_mask=attention_mask,  pad_token_id=pad_token_id,
                            penalty_alpha=penalty_alpha,num_return_sequences=2,do_sample=True,
                            max_length=max_length)
    return tokenizer.decode(output[0], skip_special_tokens=True)
```

Figure 12: The program code snippet for Contrastive Search

```
prompt = "Can you explain the concept of artificial intelligence?"
print("Contrastive Search (Vanilla):", contrastive_search_pipeline(prompt))

Contrastive Search (Vanilla): Can you explain the concept of artificial intelligence?

Well, the concept's basically this idea that if there really is a general ability to perform tasks (or
learn) then in some sense such an ability as to read the user's memories might actually help him, and
thus, perhaps at that level of performance, so much better than the human brain, perhaps to have the
ability to see things, rather than doing them by way of a computer. That's a big, complicated idea, but
what it means essentially is that you're not actually just seeing things; you're interacting with them,
which is also probably the most profound and profound of all human abilities. So as long as that ability
is useful when you have it, then it's pretty much the sort of ability there is. So that's what really
makes its existence possible, that it would be hard to imagine, on Earth, anything better.
```

Figure 13: The first specimen output text generated by Contrastive Search

```
prompt = "Can you explain the impact of social media on human relationships"
print("Contrastive Search (Vanilla):", contrastive_search(prompt))

Can you explain the impact of social media on human relationships?

It's always been a part of my life. As an adult, I've seen the first five years of my life where everyone
in the room was talking about it and my sister was looking at me. She said to me, "You know what? This is
so cool." But at the time, of course people would be like "Yeah, there is social media," and we'd be in a
conversation and say, "What happens when everyone just turns off and sees other people staring at them?"
And there would be that moment where somebody could say to another person, "Oh I want to see that other
person get my boyfriend?" And that's a huge impact on people. It's such a great way to look at what we
get wrong when others are looking and it makes us more relaxed than we are now. People will have a lot of
friends now and they want to give you the biggest hug you possibly can get.
```

Figure 14: The second specimen output text generated by Contrastive Search

```
def typical_sampling_search(prompt, max_length=200, no_repeat_ngram_size=2,
                            top_k=40,typical_p=0.95):
    input_ids = tokenizer.encode(prompt, return_tensors="pt").to(device)
    attention_mask = torch.ones_like(input_ids)
    output = model.generate(
    input_ids=input_ids,attention_mask=attention_mask,pad_token_id=tokenizer.eos_token_id,
        max_length=max_length,no_repeat_ngram_size=no_repeat_ngram_size,
        do_sample=True,top_k=top_k,typical_p=typical_p
    )
    return tokenizer.decode(output[0], skip_special_tokens=True)
```

Figure 15: The program code snippet for Locally Typical Sampling

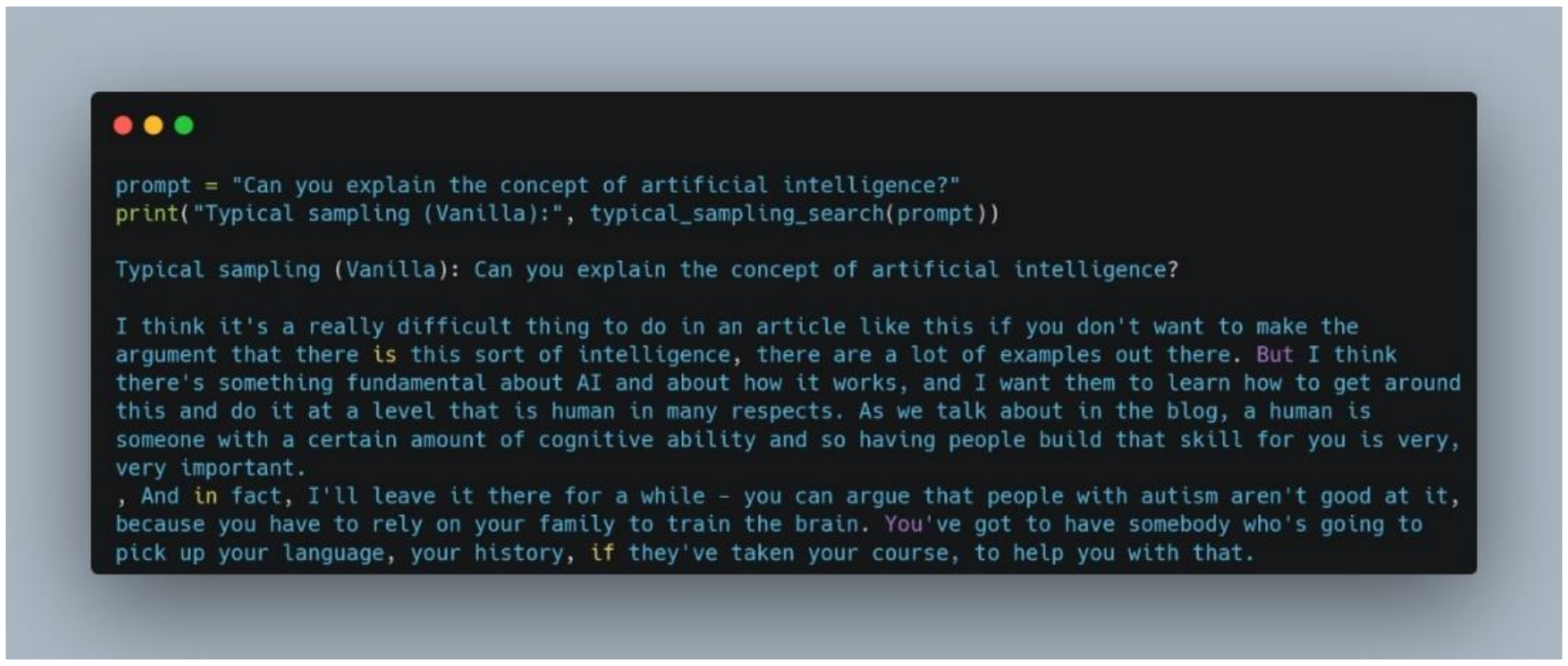

Figure 16: The first specimen output text generated by Locally Typical Sampling

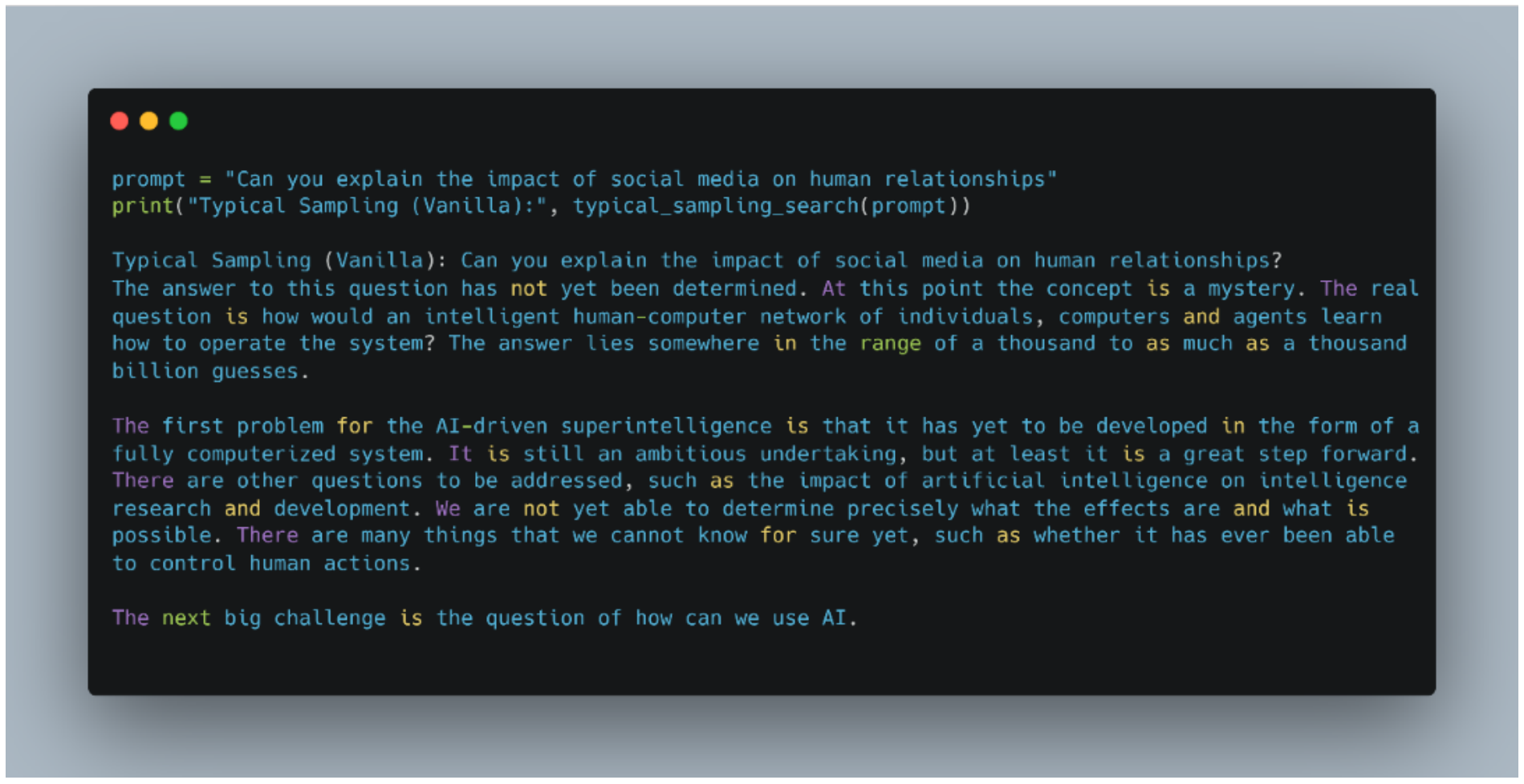

Figure 17: The second specimen output text generated by Locally Typical Sampling

A careful look at the outputs generated by the different text-generators makes it clear that the texts generated by Locally Typical Sampling and Contrastive Search are much closer and similar to texts generated by humans while the texts generated by the other models have imperfections. In Section 4.2, various metrics used in evaluating the text generators are discussed and a detailed comparative analysis of the performance of these methods is presented.

## 5.2 Performance Results of the Text-Generators

In this Section, several metrics for evaluating the performance of text generators are first presented. These metrics are used to evaluate the several qualities of text generated by the text generators.

Based on these metrics, a detailed comparative analysis of the text generators is presented subsequently.

First, several metrics are discussed concerning their utilities and methods of computation. These metrics are as follows:

**Perplexity:** Perplexity measures how well a language model predicts a sample of text. Lower perplexity indicates better performance, as it suggests that the model is more confident and accurate in predicting the next token in a sequence. Perplexity is calculated using the probability distribution generated by the language model. It's essentially the exponential of the average negative log-likelihood of each word in the test dataset. Mathematically, it is represented as in (7).

$$Perplexity = \exp\left(\frac{1}{N}\sum_{i=1}^{N} -logP(w_i\,|w_{i-1}, w_{i-2} \dots \dots .\, w_1)\right) \quad (7)$$

where $N$ is the total number of words in the test dataset, $w_i$ represents the $i$-th word in the test dataset, $P(w_i|w_{i-1}, w_{i-2}, \dots \dots w_1)$ is the probability assigned by the language model to the $i$-th word given the preceding words.

In simpler terms, perplexity can be interpreted as the average number of choices the model has for the word. Lower perplexity indicates that the model is more confident and has fewer choices, while higher perplexity indicates more uncertainty and a wider range of choices for the next word. Therefore, a lower perplexity is generally desirable in language modeling tasks

**BLEU Score:** The Bilingual Evaluation Understudy (BLEU) score is a metric used to evaluate the quality of machine-generated text, particularly in machine translation tasks, but it is also be applied to text generation tasks more broadly. In the context of text generation, particularly for tasks like language generation or summarization, BLEU score measures the similarity between the generated text and one or more reference texts (usually human-written). It quantifies how much the generated text overlaps with the reference texts in terms of *n*-grams (sequences of *n* consecutive words). The value of *n* is usually taken as 4. BLEU score works as follows:

- *N*-gram overlap: BLEU computes the precision of n-grams (typically unigrams, bigrams, trigrams, etc.) in the generated text compared to the reference text(s). Precision is the ratio of the number of *n*-grams in the generated text that appear in the reference text(s) to the total number of *n*-grams in the generated text
- Brevity penalty: BLEU also penalizes shorter translated or generated texts, favoring translations or generations that are close in length to the reference text(s). This is to ensure that the generated text is not overly concise.
- Combination: BLEU combines precision scores for different *n*-gram orders using a weighted geometric mean. This means that BLEU gives higher scores to translations or generations that have good matches for multiple lengths of *n*-grams.

The BLEU score ranges from 0 to 1, where 1 indicates a perfect match between the generated text and the reference text(s). Higher BLEU scores generally indicate better quality of the generated text, although it's important to note that BLEU has its limitations and may not always correlate perfectly with human judgment.

In summary, for text generators, a higher BLEU score suggests that the generated text is more similar to the reference text(s) and thus of higher quality according to this metric., However, BLEU scores should be used in conjunction with other metrics and human evaluation for a more comprehensive assessment.

**ROUGE Score:** Recall-Oriented Understudy for Gisting Evaluation (ROUGE) is a set of metrics used for evaluating the quality of machine-generated text summaries by comparing them to reference summaries created by humans. ROUGE measures the overlap of n-grams, word sequences, and word usage between the generated summary and the reference summary. There are several variants of ROUGE, including ROUGE-N, ROUGE-L, and ROUGE-W. Each variant focuses on different aspects of overlap between the generated summary and the reference summary:

- ROUGE-N (*N*-gram overlap): Measures the overlap of n-grams (contiguous sequences of n words) between the generated summary and the reference summary. ROUGE-N typically considers unigrams, bigrams, trigrams, etc., to capture different levels of phrase similarity.
- ROUGE-L (Longest Common Subsequence): Measures the longest common subsequence of words between the generated summary and the reference summary. It accounts for the ordering of words in the summaries and gives credit for partial matches.
- ROUGE-W (Weighted Overlap): Similar to ROUGE-N, but it assigns weights to each n-gram based on its frequency of occurrence in the reference summary. This helps mitigate the bias towards common n-grams.

The ROUGE scores range from 0 to 1, with higher scores indicating better agreement between the generated summary and the reference summary. Like BLEU, ROUGE is widely used in natural language processing tasks, particularly in text summarization and machine translation, to objectively evaluate the quality of the generated text.

In summary, ROUGE scores provide a quantitative measure of how well a text generator's output aligns with human-written reference summaries, helping researchers and practitioners assess the effectiveness of their models and make improvements.

**Perplexity-based Metrics:** Metrics like Self-BLEU and Self-ROUGE, are variants of traditional BLEU and ROUGE metrics, respectively, that are adapted for evaluating text generation models based on perplexity scores. These metrics provide a way to assess the diversity and quality of text generated by a model by comparing it against itself rather than human-generated reference texts.

- **Self-BLEU:** It measures the diversity of text generated by a model by comparing each generated sentence against other sentences produced by the same model. Instead of using human-generated reference texts, Self-BLEU computes BLEU scores between each generated sentence and the rest of the sentences produced by the model. A lower Self-BLEU score indicates higher diversity because it suggests that the generated sentences are less similar to each other. It helps in assessing whether a model tends to produce repetitive or overly similar outputs. Self-BLEU is calculated by averaging the BLEU scores of each sentence against all other sentences or by considering a subset of sentences for comparison.

- **Self-ROUGE:** It is similar to Self-BLEU but is based on the ROUGE metric, which is commonly used for evaluating text summarization tasks. Instead of using BLEU, Self-ROUGE computes ROUGE scores between each generated sentence and the rest of the sentences produced by the model. Just like Self-BLEU, a lower Self-ROUGE score indicates higher diversity among the generated sentences. Self-ROUGE helps in assessing the coherence and diversity of generated text summaries or outputs. It can also be calculated by averaging the ROUGE scores of each sentence against all other sentences or by considering a subset of sentences for comparison.

In summary, Perplexity-based metrics such as Self-BLEU and Self-ROUGE provide valuable insights into the diversity and quality of text generated by a model without relying on external reference texts. These metrics are particularly useful for assessing the diversity, coherence, and repetitiveness of generated text, which are essential aspects of evaluating text generation models

**Diversity Metrics:** Metrics such as Distinct-n and Entropy are used to measure different aspects of the diversity of text generated by language models or text generation systems. Higher diversity indicates a wider range of vocabulary and more varied output.

- **Distinct-*n***: Distinct-*n* measures the diversity of vocabulary used in the generated text by counting the number of unique *n*-grams (sequences of *n* words) in the generated text. It's calculated by dividing the total count of unique *n*-grams by the total count of *n*-grams in the generated text. Distinct-*n* can be computed for different values of *n* (e.g., unigrams, bigrams, trigrams, etc.), providing insights into the diversity of single words, pairs of words, or longer phrases used in the generated text. A higher distinct-*n* score indicates greater diversity in the vocabulary used, suggesting a richer and more varied expression in the generated text.

- **Entropy:** Entropy is a measure of uncertainty or randomness in the distribution of words or tokens in the generated text. In the context of text generation, entropy measures how evenly the probability mass is distributed across different words or tokens. It's calculated using the probability distribution of words or tokens in the generated text, typically based on their frequency of occurrence. Higher entropy values indicate a more uniform distribution of words or tokens, suggesting greater diversity or randomness in the generated text. Conversely, lower entropy values indicate a more predictable distribution, where certain words or tokens are favored over others, potentially indicating less diversity in the generated text.

Both distinct-*n* and entropy are useful diversity metrics for evaluating text generation systems, providing complementary insights into the richness, variety, and randomness of the generated text. They help in assessing and comparing the diversity of text generated by different models.

**Coherence Metrics:** Coherence metrics in text generation evaluate how well the generated text flows logically and maintains a coherent structure. These metrics assess the overall coherence of the generated text, including the organization of ideas, coherence within paragraphs, and the logical progression of the narrative. Some common coherence metrics are as follows.

- **Connectivity:** Connectivity metrics evaluate the flow of ideas and the coherence of transitions between sentences or paragraphs in the generated text. These metrics may involve analyzing the use of cohesive devices such as pronouns, conjunctions, transitional phrases, and discourse markers to establish logical connections between sentences and paragraphs.

- **Cohesion:** Cohesion metrics assess the internal consistency and coherence within individual sentences or paragraphs. They examine how well the components of a sentence or paragraph are linked together through grammatical and semantic relationships, including reference resolution, lexical cohesion, and syntactic coherence.

- **Topic Coherence:** Topic coherence metrics evaluate the relevance and consistency of the generated text for the given topic or prompt. They assess whether the generated content aligns with the main theme or subject matter and maintains coherence with the intended topic throughout the text.

- **Entity Consistency:** Entity consistency metrics examine the consistency and coherence in the representation of entities (e.g., characters, locations, events) across the generated text. They assess whether entities are introduced and referred to consistently, without contradictions or inconsistencies, throughout the narrative.

- **Semantic Coherence:** Semantic coherence metrics evaluate the logical consistency and coherence of meaning conveyed by the generated text. They assess whether the semantic relationships between concepts, events, and actions are coherent and logically connected, without ambiguities or contradictions.

- **Readability:** Readability metrics assess how easily the generated text can be understood and comprehended by readers. These metrics may consider factors such as sentence structure, vocabulary complexity, readability scores (e.g., Flesch-Kincaid Grade Level), and syntactic clarity to evaluate the overall coherence and accessibility of the text.

Coherence metrics play a crucial role in evaluating the quality and naturalness of generated text, as they capture aspects related to logical coherence, topical relevance, and readability, which are essential for producing coherent and engaging narratives. These metrics are used in combination with other evaluation criteria to provide a comprehensive assessment of text generation systems.

**Relevance Metrics:** Relevance metrics for text generators assess how well the generated text aligns with the intended task, prompt, or context. These metrics evaluate the topical relevance, informativeness, and appropriateness of the generated text for the given input or target. Some of the common relevance metrics used for evaluating text generators are as follows.

- **Prompt Adherence:** Prompt adherence metrics evaluate the extent to which the generated text adheres to the provided prompt or task instructions. They assess whether the content generated by the model addresses the specific requirements, constraints, or prompts provided as input.

- **Topic Relevance:** Topic relevance metrics assess the relevance of the generated text to the specified topic or subject matter. They evaluate whether the content generated by the model is related to the main theme or topic provided as input, ensuring that the generated text stays on topic.

- **Information Completeness:** Information completeness metrics evaluate the extent to which the generated text provides comprehensive and relevant information related to the input or task. They assess whether the generated content adequately covers the key aspects, details, or points relevant to the given topic or prompt.

- **Factuality:** Factuality metrics assess the accuracy and truthfulness of the information presented in the generated text. They evaluate whether the statements, claims, or assertions made in the generated content are factually accurate and supported by evidence or sources.

- **Relevance to User Intent:** Relevance to user intent metrics evaluate how well the generated text fulfills the intended needs, goals, or preferences of the target audience or user. They assess whether the content generated by the model effectively addresses the user's queries, requests, or objectives, ensuring relevance and usefulness.

- **Contextual Coherence:** Contextual coherence metrics evaluate the relevance and coherence of the generated text within the broader context, including situational, cultural, or domain-specific contexts. They assess whether the content generated by the model takes into account contextual cues, background information, or contextual constraints to produce relevant and contextually appropriate responses.

The relevance metrics help in evaluating the quality, appropriateness, and effectiveness of text generation systems in producing relevant and informative content. They provide valuable insights into the performance and suitability of text generators for specific tasks, applications, or use cases.

**Grammar and Syntax Metrics:** Grammar and syntax metrics for text generators evaluate the grammatical correctness, syntactic structure, and fluency of the generated text. These metrics help assess how well the text conforms to the rules of grammar and syntax, ensuring readability and linguistic correctness. Following are some common grammar and syntax metrics for text generators.

- **Grammar Error Rate:** Grammar error rate metrics quantify the percentage of sentences or tokens in the generated text that contain grammatical errors. They identify errors such as subject-verb agreement errors, tense inconsistencies, punctuation errors, pronoun errors, and other syntactic inaccuracies.

- **Syntactic Complexity:** Syntactic complexity metrics assess the complexity of sentence structures and syntactic constructions used in the generated text. They measure features such as sentence length, clause density, subordination, coordination, and syntactic variety to evaluate the richness and sophistication of the syntactic structures.

- **POS Tagging Accuracy:** Part-of-speech (POS) tagging accuracy metrics evaluate the accuracy of the POS tags assigned to words in the generated text. They compare the POS tags predicted by the model with the gold-standard POS tags to measure the model's ability to capture the syntactic properties and grammatical functions of words in context.

- **Syntax Tree Alignment:** Syntax tree alignment metrics assess the alignment between the syntactic structures of the generated text and the syntactic structures of reference texts or gold-standard syntactic annotations. They compare the parse trees or syntactic dependencies generated by the model with the parse trees or dependencies of human-written texts to evaluate syntactic fidelity and alignment.

- **Dependency Parsing Accuracy:** Dependency parsing accuracy metrics evaluate the accuracy of dependency relations predicted by the model between words in the generated text. They compare the predicted dependency relations with the gold-standard dependency relations to assess the model's ability to capture syntactic dependencies and relationships.

- **Fluency:** Fluency metrics assess the overall smoothness, coherence, and naturalness of the generated text in terms of grammar and syntax. They evaluate factors such as sentence structure, word order, phrasing, and idiomatic expressions to measure the fluency and naturalness of the generated language. These grammar and syntax metrics provide quantitative measures of the grammatical correctness, syntactic complexity, and fluency of text generated by language models or text generation systems. By evaluating these aspects, researchers and practitioners can assess the linguistic quality and syntactic accuracy of text generators for various applications and domains.

The grammar and syntax metrics provide quantitative measures of the grammatical correctness, syntactic complexity, and fluency of text generated by language models or text generation systems. By evaluating these aspects, it is possible to assess the linguistic quality and syntactic accuracy of text generators for various applications and domains.

**Semantic Similarity Metrics:** Semantic similarity metrics for text generation assess how closely the generated text matches the meaning, semantics, and underlying concepts of the target or reference text. These metrics evaluate the semantic relatedness, similarity, and coherence between the generated text and the reference text based on their semantic representations. Some common semantic similarity metrics used for evaluating text generation are as follows.

- **Word Embedding Similarity:** Word embedding similarity metrics compute the similarity between word embeddings of words in the generated text and the reference text. They use pre-trained word embeddings (e.g., Word2Vec, GloVe, FastText) to represent words as dense vectors in a high-dimensional space and calculate cosine similarity, Euclidean distance, or other similarity measures between word vectors.

- **Sentence Embedding Similarity:** Sentence embedding similarity metrics compute the similarity between sentence embeddings of sentences in the generated text and the reference text. They use methods such as averaging word embeddings, Doc2Vec, or transformer-based sentence encoders (e.g., BERT, RoBERTa) to generate fixed-length representations of sentences and calculate similarity measures between sentence embeddings.

- **Semantic Textual Similarity:** Semantic Textual Similarity metrics evaluate the degree of semantic similarity or relatedness between pairs of sentences or texts. They typically use human-labeled datasets with graded similarity scores to train models that predict the semantic similarity between sentences based on various features, including word overlap, syntactic similarity, and semantic relatedness (Waghela et al., 2024).

- **BERTScore:** BERTScore computes the similarity between the contextual embeddings of tokens in the generated text and the reference text using BERT-based embeddings. It leverages contextual embeddings from BERT or other transformer-based models to capture the contextual similarity and semantic equivalence between tokens in the generated and reference texts.

- **Meaning Equivalence:** Meaning equivalence metrics assess whether the generated text conveys the same underlying meaning, concepts, or information as the reference text. They compare the semantic representations or conceptual content of the generated and reference texts to evaluate their equivalence regarding meaning and semantic content.

- **Distributional Semantic Models (DSMs):** Distributional Semantic Models capture the distributional properties of words or phrases based on their co-occurrence patterns in a large corpus. They compute similarity or relatedness scores between words or phrases in the generated and reference texts using distributional similarity measures, such as cosine similarity or Jaccard similarity.

The semantic similarity metrics provide quantitative measures of the semantic relatedness, coherence, and equivalence between the generated text and the reference text. By evaluating these aspects, it is possible to evaluate the semantic quality and fidelity of text generated by language models or text generation systems.

**Human Evaluation:** Ultimately, human judgment remains a crucial aspect of evaluating text generation quality. Human evaluations for text-generating systems involve collecting feedback and assessments from human evaluators to measure various aspects of the generated text's quality, including coherence, relevance, fluency, and overall satisfaction. Human evaluations provide valuable insights into the subjective aspects of text quality that may not be captured by automated metrics. Following are some common methods used for human evaluations of text-generating systems.

- **Human Judgments:** Human evaluators read and assess the generated text based on predefined criteria, such as coherence, relevance, fluency, grammaticality, and overall quality. They provide subjective judgments, ratings, or scores for each aspect, often using Likert scales or qualitative descriptors (e.g., excellent, good, fair, poor).

- **Pairwise Comparisons:** Human evaluators compare pairs of generated texts and select the one that they consider better according to specified criteria. Pairwise comparisons can be used to assess the relative quality or preference between different text generation models, variations, or settings.

- **Ranking:** Human evaluators rank multiple generated texts in order of preference or quality based on specified criteria. Ranking evaluations provide insights into the relative performance of different text generation systems or approaches.

- **Preference Tasks:** Human evaluators express their preferences for one generated text over another or specific attributes of the generated text (e.g., coherence, relevance, creativity). Preference tests help identify which aspects of the generated text are most important to users or which text generation systems are preferred overall.

- **Annotation Tasks:** Human evaluators annotate specific aspects of the generated text, such as identifying errors, highlighting relevant information, or providing feedback on coherence and clarity. Annotation tasks can involve manual correction of errors, highlighting key passages, or providing comments and suggestions for improvement.

- **Qualitative Feedback:** Human evaluators provide qualitative feedback, comments, and insights on the strengths, weaknesses, and overall impressions of the generated text. Qualitative feedback can include detailed analyses, observations, and suggestions for enhancing the quality and usability of text generation systems.

- **User Studies:** User studies involve collecting feedback and evaluations from end-users who interact with the generated text in real-world scenarios or applications. User studies assess user satisfaction, usability, and effectiveness of text generation systems in meeting user needs and preferences.

Human evaluations complement automated metrics by providing subjective assessments and qualitative insights into the perceived quality, usability, and effectiveness of text-generating systems. By incorporating human judgments and feedback, researchers and practitioners can gain a deeper understanding of user preferences, identify areas for improvement, and guide the development of more effective text-generation models and applications.

Based on the metrics mentioned in Section 5. 2, a comprehensive comparative analysis is made for the text-generation methods presented in Section 4. The results of the comparative analysis presented in the following are based on the observations of the executions of the programs illustrated in Section 5.1 and also information available in the literature on Generative AI-based text generation methods and Large Language Models (LLMs).

Table 1: Comparative Analysis of the Perplexity Metrics

| Generation Method | Performance |
|---|---|
| Greedy Search | Moderate to High |
| Beam Search | Similar to Greedy but with potentially lower |
| Top-K Sampling | Moderate |
| Top-P Sampling | Similar to Top-K with potentially lower |
| Contrastive Search | Moderate to Low |
| Locally Typical Sampling | Moderate to Low |

Table 1 presents the performance of the text-generation methods on the Perplexity metrics. Greedy Search has a moderate to high perplexity due to its tendency to favor locally optimal choices without considering the broader context. The performance of Beam Search is similar to the Greedy Search, but with potentially lower perplexity as it explores multiple hypotheses simultaneously. Top-K Sampling has a moderate perplexity, balancing between exploration and exploitation, with the ability to produce diverse outputs. Top-P Sampling performs similarly to Top-K, with a potentially lower perplexity by dynamically adjusting the probability distribution. Contrastive Search has moderate to low perplexity, encouraging the model to produce diverse and high-quality outputs. Locally Typical Sampling has moderate to low perplexity, focusing on generating text that is locally typical within a given context.

Table 2: Comparative Analysis of the BLEU Score Metrics

| **Generation Method** | **Performance** |
|---|---|
| Greedy Search | Low |
| Beam Search | Moderate |
| Top-K Sampling | Moderate to High |
| Top-P Sampling | Similar to Top-K, possibly Higher |
| Contrastive Search | Moderate to High |
| Locally Typical Sampling | Moderate to High |

The performance results of the methods on BLEU Score metrics are presented in Table 2. Greedy Search yields a low BLEU score due to its tendency to produce repetitive and less diverse outputs. Beam Search produces moderate BLEU scores but often suffers from generating overly conservative and repetitive text. Top-K Sampling most often yields moderate to high BLEU scores, as it is capable of producing diverse outputs while maintaining coherence with the reference texts. Top-P Sampling produces results that are similar to Top-K Sampling, with potentially higher BLEU scores by dynamically adjusting the sampling strategy. Contrastive Search produces moderate to high BLEU scores, encouraging the model to produce diverse and relevant outputs.

Locally Typical Sampling yields moderate to high BLEU scores, focusing on generating text that aligns with typical language patterns.

Table 3: Comparative Analysis of the ROUGE Score Metrics

| Generation Method | Performance |
|---|---|
| Greedy Search | Low |
| Beam Search | Moderate |
| Top-K Sampling | Moderate to High |
| Top-P Sampling | Similar to Top-K, possibly Higher |
| Contrastive Search | Moderate to High |
| Locally Typical Sampling | Moderate to High |

Table 3 presents the performance of the text-generation methods on the ROUGE Score metrics. Greedy Search typically yields a low ROUGE score due to its tendency to produce repetitive and less diverse outputs. Beam Search yields a moderate ROUGE score, but often lacks diversity and may not capture nuanced aspects of the reference texts. Top-K Sampling produces moderate to high ROUGE scores since it is capable of producing diverse outputs while maintaining relevance to the reference texts. Top-P Sampling yields similar ROUGE scores as Top-K sampling, having the potential of producing higher scores by dynamically adjusting the sampling strategy. Contrastive Search yields moderate to high ROUGE scores, encouraging the model to produce diverse and relevant outputs. Locally Typical Sampling produces moderate to high ROUGE scores, focusing on generating coherent and contextually relevant text.

Table 4: Comparative Analysis of the Diversity Metrics

| Generation Method | Performance |
|---|---|
| Greedy Search | Low |
| Beam Search | Moderate |
| Top-K Sampling | High |
| Top-P Sampling | Similar to Top-P possibly Higher |
| Contrastive Search | High |
| Locally Typical Sampling | Moderate to High |

Table 4 presents the performance of the text-generation methods on the Diversity Score metrics. Greedy Search typically yields a low diversity, often producing repetitive and predictable outputs. Beam Search produces a moderate diversity but its ability is limited compared to stochastic sampling methods. Top-K Sampling produces a high diversity as it is capable of generating a wide range of outputs with varied vocabulary. Top-P Sampling produces similar results as Top-K Sampling, with potentially higher diversity by dynamically adjusting the sampling strategy. Contrastive Search yields higher diversity values, as it encourages the model to explore different hypotheses and produce diverse outputs. Locally Typical Sampling produces a moderate to high diversity, focusing on generating text that exhibits typical language patterns while maintaining diversity.

Table 5: Comparative Analysis of the Coherence Metrics

| **Generation Method** | **Performance** |
|---|---|
| Greedy Search | Low to Moderate |
| Beam Search | Similar to Greedy possibly Higher |
| Top-K Sampling | Moderate to High |
| Top-P Sampling | Similar to Top-K possibly Higher |
| Contrastive Search | Moderate to High |
| Locally Typical Sampling | High |

The performance results of the methods on Coherence metrics are presented in Table 5. Greedy Search typically yields moderate coherence within local contexts but lacks overall coherence in longer texts. Beam Search has a similar performance as Greedy Search, with slightly better overall coherence due to the exploration of multiple hypotheses. Top-K Sampling produces moderate to high coherence and is capable of producing contextually relevant and fluent text. Top-P Sampling performs similarly to Top-K Sampling, with potentially higher coherence by dynamically adjusting the sampling strategy. Contrastive Search has moderate to high coherence, encouraging the model to produce coherent and contextually relevant outputs. Locally Typical Sampling yields high coherence, focusing on generating text that flows logically and cohesively within a given context.

The performance results of the text-generation methods on Relevance metrics are presented in Table 6. Greedy Search yields moderate relevance to the given prompt or context but may lack diversity most often. Beam Search performs similarly to Greedy search, with slightly better relevance due to the exploration of multiple hypotheses. Top-K Sampling produces high relevance since it is capable of producing contextually relevant and diverse outputs. Top-P Sampling performs similarly to Top-K Sampling, with potentially higher relevance by dynamically adjusting the sampling strategy. Contrastive Search yields high relevance, as it encourages the model to produce outputs that are semantically similar to the reference texts. Locally Typical Sampling produces high relevance, focusing on generating text that aligns with typical language patterns within a given context.

Table 6: Comparative Analysis of the Relevance Metrics

| **Generation Method** | **Performance** |
|---|---|
| Greedy Search | Moderate |
| Beam Search | Similar to Greedy, possibly a little better |
| Top-K Sampling | High |
| Top-P Sampling | Similar to Top-K, possibly Higher |
| Contrastive Search | High |
| Locally Typical Sampling | High |

The performance results of the methods on Grammar & Syntax metrics are presented in Table 7. Due to its lack of exploration, Greedy Search may struggle to correct errors or inconsistencies in the generated text, resulting in poorer performance on grammar and syntax metrics. While Beam Search can mitigate some grammatical errors compared to Greedy Search, it may still produce suboptimal solutions and exhibit repetition or lack of diversity, which can impact its performance on grammar and syntax metrics. Top-K and Top-P Sampling introduce randomness and diversity into the generated text, which can help in exploring different syntactic structures and reducing repetitive patterns. However, they may also introduce grammatical errors or syntactic inconsistencies, particularly when sampling from low-probability tokens or when the sampling distribution is skewed. The performance of Top-K and Top-P Sampling on grammar and syntax may vary depending on the chosen values of *k* and *p* and the specific characteristics of the text-generation task. Contrastive Search optimizes for semantic similarity but may indirectly improve grammar and syntax by encouraging the model to produce text that aligns well with the reference text. By penalizing text that diverges too much from the reference, Contrastive Search may promote grammatically correct and syntactically coherent outputs. However, its performance on grammar and syntax metrics may still depend on the quality of the reference text and the effectiveness of the contrastive objective function. Locally Typical Sampling biases the sampling distribution towards tokens that are typical for the local context. By favoring tokens that are likely to occur in the current context, Locally Typical Sampling produces text that adheres more closely to grammar and syntactic conventions.

Table 7: Comparative Analysis of the Grammar & Syntax Metrics

| **Generation Method** | **Performance** |
|---|---|
| Greedy Search | Poor most often |
| Beam Search | Better than Greedy but Poor most often |
| Top-K Sampling | Varies depending on the values of k. Most often, introduces errors or syntactic inconsistencies |
| Top-P Sampling | Most often similar to Top-K, at times better |
| Contrastive Search | Most often produces grammatically correct and syntactically coherent output. However, its performance depends on the quality of reference texts and the effectiveness of the objective function |
| Locally Typical Sampling | Generates texts which grammatically and syntactically correct all the time. Performs the best among all these methods |

The performance results of the methods on Grammar & Syntax metrics are presented in Table 7. Due to its lack of exploration, Greedy Search may struggle to correct errors or inconsistencies in the generated text, resulting in poorer performance on grammar and syntax metrics. While Beam Search can mitigate some grammatical errors compared to Greedy Search, it may still produce suboptimal solutions and exhibit repetition or lack of diversity, which can impact its performance on grammar and syntax metrics. Top-K and Top-P Sampling introduce randomness and diversity into the generated text, which can help in exploring different syntactic structures and reducing repetitive patterns. However, they may also introduce grammatical errors or syntactic inconsistencies, particularly when sampling from low-probability tokens or when the sampling distribution is skewed. The performance of Top-K and Top-P Sampling on grammar and syntax may vary depending on the chosen values of $k$ and $p$ and the specific characteristics of the text-generation task. Contrastive Search optimizes for semantic similarity but may indirectly improve grammar and syntax by encouraging the model to produce text that aligns well with the reference text. By penalizing text that diverges too much from the reference, Contrastive Search may promote grammatically correct and syntactically coherent outputs. However, its performance on grammar and syntax metrics may still depend on the quality of the reference text and the effectiveness of the contrastive objective function. Locally Typical Sampling biases the sampling distribution towards tokens that are typical for the local context. By favoring tokens that are likely to occur in the current context, Locally Typical Sampling produces text that adheres more closely to grammar and syntactic conventions.

Table 8: Comparative Analysis of the Semantic Similarity Metrics

| **Generation Method** | **Performance** |
|---|---|
| Greedy Search | Poor in general |
| Beam Search | Better than Greedy but poor in general |
| Top-K Sampling | High Diversity, may lack in semantics at times due to the introduction of noise and irrelevant content |
| Top-P Sampling | Similar to Top-K, sometimes Higher |
| Contrastive Search | High most often |
| Locally Typical Sampling | Highest among all these methods |

Finally, the performance results of the methods on Semantic Similarity metrics are presented in Table 8. Greedy Search tends to produce deterministic outputs that may lack diversity and fail to capture the nuanced semantic relationships present in the reference text. As a result, it may perform relatively poorly on semantic similarity metrics compared to other algorithms that prioritize diversity and exploration. Beam Search expands multiple candidate sequences but may still suffer from suboptimal solutions and lack of diversity, especially with a narrow beam width. While it may perform better than Greedy Search due to the consideration of multiple candidates, it may still struggle to produce highly semantically similar outputs compared to more flexible sampling methods. Top-K and Top-P Sampling introduce randomness and diversity into the generated text, allowing for a wider exploration of the output space. While they may produce more diverse outputs compared to Greedy Search and Beam Search, they can also introduce noise and irrelevant content, which may adversely affect semantic similarity metrics. Contrastive Search explicitly optimizes for semantic similarity by encouraging the generated text to be similar to a reference text while being dissimilar to negative examples. It may perform relatively well on semantic similarity metrics sometimes as it directly targets semantic alignment between the generated text and the reference text. Locally Typical Sampling performs the best on semantic similarity metrics, particularly where maintaining local coherence and relevance is important.

Human evaluation is crucial for assessing subjective aspects of text quality, such as creativity, coherence, and overall fluency. While automated metrics provide valuable insights, human judgment remains indispensable for evaluating the nuanced aspects of generated text.

**Summary:** In summary, each text generation method has its own strengths and weaknesses across different evaluation metrics. Greedy search and beam search offer simplicity and computational efficiency but may produce less diverse and repetitive outputs. Stochastic sampling methods like top-k sampling and top-p sampling prioritize diversity and relevance, while contrastive search and locally typical sampling focus on producing diverse, coherent, and contextually relevant outputs. The choice of method depends on the specific requirements of the text generation task and the desired trade-offs between coherence, diversity, and relevance.

# 6. Conclusion and Future Work

In this work, we delved into the realm of automatic text generation, exploring a variety of techniques ranging from traditional deterministic approaches to more modern stochastic methods. Through our analysis of greedy search, beam search, top-k sampling, top-p sampling, contrastive searching, and locally typical searching, we have gained valuable insights into the strengths, weaknesses, and potential applications of each method.

Greedy Search serves as a fundamental baseline for text generation tasks, offering simplicity and computational efficiency. However, its myopic nature often leads to suboptimal solutions, as it tends to favor locally optimal choices without considering the broader context. While suitable for certain scenarios where speed is of the essence, greedy search falls short when confronted with the need for long-range coherence and semantic fidelity in the generated text.

Beam Search, an extension of greedy search, attempts to address some of its limitations by exploring multiple candidate sequences in parallel. By maintaining a beam of the most promising hypotheses at each step, beam search can mitigate the issue of premature convergence to suboptimal solutions. Nevertheless, it often struggles with producing diverse and creative outputs, as it tends to favor well-trodden paths through the search space, leading to repetitive and unimaginative text generation.

Top-k Sampling and Top-p Sampling represent a departure from the deterministic nature of greedy and beam search algorithms, introducing stochasticity into the generation process. By sampling from the top-k or top-p most likely tokens at each step, these methods allow for a more diverse range of outputs, thereby encouraging creativity and exploration. However, they also present challenges in balancing between exploration and exploitation, as excessively large or small values of k or p can lead to either overly diverse or overly repetitive outputs, respectively.

Contrastive Searching represents a novel approach to text generation, drawing inspiration from contrastive learning techniques in machine learning. By training a discriminator to distinguish between real and generated sequences, contrastive searching encourages the model to produce diverse and high-quality outputs. This method shows promise in addressing issues such as mode collapse and lack of diversity in generated text. However, it comes with its own set of challenges, including the need for additional computational resources and careful tuning of hyperparameters.

Locally Typical Searching takes a different approach to text generation by focusing on generating text that is locally typical within a given context. By leveraging knowledge about typical language patterns and syntactic structures, this method aims to produce more coherent and contextually relevant outputs. While effective in capturing short-range dependencies and ensuring fluency, locally typical searching may struggle with generating novel or creative content, as it tends to prioritize conformity to established norms over exploration of alternative possibilities.

In summary, the choice of text generation technique depends on a multitude of factors, including the specific requirements of the task, the computational resources available, and the desired characteristics of the generated text. While each method has its own set of advantages and limitations, the field of automatic text generation continues to evolve rapidly, with researchers exploring innovative approaches and hybrid strategies to overcome existing challenges.

Looking ahead, future research directions in automatic text generation may involve:

Algorithmic Innovations: Developing novel algorithms and architectures that address the shortcomings of existing methods, such as improving the balance between exploration and exploitation, enhancing sample diversity, and capturing long-range dependencies more effectively.

Interpretability and Controllability: Investigating methods for making generated text more interpretable and controllable, allowing users to influence the style, tone, and content of the generated output more explicitly.

Ethical Considerations: Examining the ethical implications of AI-generated content, including issues related to bias, fairness, misinformation, and privacy, and developing guidelines and frameworks for responsible deployment of text generation models.

Multimodal Text Generation: Exploring approaches for generating text in conjunction with other modalities, such as images, audio, and video, to enable more immersive and interactive experiences in applications such as storytelling, content creation, and virtual environments.

Real-World Applications: Investigating the practical applications of automatic text generation in various domains, including journalism, education, healthcare, customer service, and creative industries, and evaluating the impact of text generation technologies on society and culture.

By addressing these research challenges and opportunities, we can unlock the full potential of automatic text generation technology, paving the way for new forms of human-computer interaction, communication, and creative expression in the digital age.

In conclusion, automatic text generation represents a fascinating and rapidly evolving field at the intersection of artificial intelligence, natural language processing, and human-computer interaction. Through continuous innovation, collaboration, and interdisciplinary research, we can harness the power of text generation to enrich human communication, foster creativity, and advance our understanding of language and culture in the 21st century and beyond.